%% file: ms.tex
\documentclass{article}

\usepackage{arxiv}

\usepackage[utf8]{inputenc} 
\usepackage[T1]{fontenc}    
\usepackage{hyperref}       
\usepackage{url}            
\usepackage{booktabs}       
\usepackage{amsfonts}       
\usepackage{nicefrac}       
\usepackage{microtype}      
\usepackage{lipsum}		
\usepackage{graphicx}
\usepackage{natbib}
\usepackage{doi}
\usepackage{acronym}
\usepackage{amsmath}
\usepackage{multirow}
\usepackage{subfig}
\usepackage{graphicx}
\usepackage{tikz}
\usepackage{caption}
\usetikzlibrary{calc}
\usetikzlibrary {positioning}
\usepgfmodule {nonlineartransformations}
\usetikzlibrary {curvilinear}
\newcommand{\uproman}[1]{\uppercase\expandafter{\romannumeral#1}}

\title{A Reinforcement Learning Approach for Scheduling Problems With Improved Generalization Through Order Swapping}

\author{{Deepak Vivekanandan} \thanks{main author} \\
Scaliro GmbH\\
\texttt{deepak.vivekanandan@scaliro.de}
 \And {Samuel Wirth} \\
    Rosenheim University of Applied Sciences \\
	\texttt{samuel.wirth@th-rosenheim.de} \\
 \And {Patrick Karlbauer} \\
    Rosenheim University of Applied Sciences\\
	\texttt{patrick.karlbauer@th-rosenheim.de} \\
 \And {Noah Klarmann} \\
    Rosenheim University of Applied Sciences\\
	\texttt{noah.klarmann@th-rosenheim.de} \\
}

\renewcommand{\headeright}{}
\renewcommand{\undertitle}{}
\renewcommand{\shorttitle}{}
\DeclareUnicodeCharacter{2061}{}

\hypersetup{
pdftitle={A partially generalized deep reinforcement learning approach to solver job shop scheduling problems},
pdfsubject={q-bio.NC, q-bio.QM},
pdfauthor={Deepak Vivekanandan},
pdfkeywords={Job Shop Problem, Reinforcement Learning, ...},
}

\begin{document}
\maketitle

\begin{abstract}
	The scheduling of production resources (such as associating jobs to machines) plays a vital role for the manufacturing industry not only for saving energy but also for increasing the overall efficiency. Among the different job scheduling problems, the \ac{jssp} is addressed in this work. \ac{jssp} falls into the category of NP-hard \ac{cop}, in which solving the problem through exhaustive search becomes unfeasible. Simple heuristics such as \ac{fifo}, \ac{lpt} and metaheuristics such as Taboo search are often adopted to solve the problem by truncating the search space. The viability of the methods becomes inefficient for large problem sizes as it is either far from the optimum or time consuming. In recent years, the research towards using \ac{drl} to solve \ac{cop}s has gained interest and has shown promising results in terms of solution quality and computational efficiency. In this work, we provide an novel approach to solve the \ac{jssp} examining the objectives generalization and solution effectiveness using \ac{drl}. In particular, we employ the \ac{ppo} algorithm that adopts the policy-gradient paradigm that is found to perform well in the constrained dispatching of jobs. We incorporated an \ac{osm} in the environment to achieve better generalized learning of the problem. The performance of the presented approach is analyzed in depth by using a set of available benchmark instances and comparing our results with the work of other groups.
\end{abstract}

\keywords{Job Shop Scheduling \and Reinforcement Learning \and Generalization}

\begin{acronym}
    \acro{jssp}[JSSP]{Job Shop Scheduling Problem}
    \acro{cop}[COP]{Combinatorial Optimization Problem}
    \acro{fifo}[FIFO]{First In, First Out}
    \acro{lpt}[LPT]{Largest Processing Time First}
    \acro{ppo}[PPO]{Proximal Policy Optimization}
    \acro{osm}[OSM]{Order Swapping Mechanism}
    \acro{drl}[DRL]{Deep Reinforcement Learning}
    \acro{ai}[AI]{Artificial Intelligence}
    \acro{or}[OR]{Operational Research}
    \acro{trpo}[TRPO]{Trust Region Policy Optimization}
    \acro{tsp}[TSP]{Travelling Salesman Problem}
    \acro{rl}[RL]{Reinforcement Learning}
    \acro{mdp}[MDP]{Markov Decision Process}
    \acro{mmdp}[MMDP]{Multi-Agent Markov Decision Process}
    \acro{ddpg}[DDPG]{Deep Deterministic Policy Gradient}
    \acro{pdr}[PDR]{Priority Dispatching Rule}
\end{acronym}

\section{Introduction}
Scheduling problems in the field of manufacturing are usually distinguished in one of the three categories: (1) \ac{jssp}, (2) flow shop, and (3) open shop. This work addresses \ac{jssp}s, which are highly challenging, of significant industrial relevance, and often used as a benchmark for testing/comparing new methodologies. In \ac{jssp}s, every job has a fixed machine sequence that has to be followed during the production of the particular product \citep{pinedo2012scheduling}. Moreover, the job shop has $n$ jobs $J_0, J_1, J_2, \dots, J_n$ that must be processed on $m$ machines with every job-machine pair having a specific processing time that is given by the problem formulation. As the number of jobs and machines increases, combinatorial possibilities quickly explode and computation time of exhaustive searches become unfeasible even for medium-sized problems. It is worth noting that \ac{cop}s are considered to fall into the class of NP-hard problems. Moreover, conventional \ac{cop}s and \ac{jssp}s are structurally different, giving rise to a challenging problem in designing an effective representation \citep{zhang2020learning}.
\newline
\newline
\noindent
\ac{drl} is a subfield of machine learning where an agent is trained based on experience that is gathered from the interaction with an uncertain environment. The agent improves its performance by maximizing a reward signal that characterizes the overall goal such as reaching the shortest makespan in a production \citep{sutton2018reinforcement}. Lately, some remarkable milestones in the field of \ac{ai} have been reached by employing \ac{drl}, such as outperforming the human in popular challenges such as the board game Go (AlphaGO \citep{silver2017mastering}) or StarCraft \uproman{2} (AlphaStar \citep{vinyals2019grandmaster}). The implementation of \ac{drl} in the field of \ac{or} has become quite significant. Several studies incorporating \ac{drl} to solve \ac{cop} have shown promising results \citep{du2021vulcan, afshar2020state}. Moreover, \ac{drl} provides a significantly faster approximation for \ac{cop}s compared to exhaustive search, metaheuristics, or other conventional heuristics.
In this paper, we propose a \ac{drl} based approach to efficiently solve the \acf{jssp}. We developed an efficient, problem-generic environment for arbitrary \ac{jssp} problems in OpenAI’s gym framework. Along with the optimal reward modelling and compact state representation of the \ac{jssp} environment, the policy parameters of the policy network were trained to approach a deterministic policy. Based on the proposed approach, the \ac{ppo} algorithm was tested by solving classical benchmark problems such as \citet{taillard1993benchmarks}, and  \citet{demirkol1998benchmarks}. The performance of our trained network is compared with state-of-the-art algorithms based on computation time and makespan (time to complete all operations).

\section{Background}
\label{sec:background}
\subsection{Job Shop Constraints}
\ac{jssp}s consist of $n$ jobs that need be processed on $m$ machines. 
Each job has a respective order in which it is to be completed with respect to the other jobs and a processing time that determines when the machine is ready to process the next job.
The total number of operations equals  $O \rightarrow n \times m$. Each operation is indicated by $O_{ij}$ and their respective processing time is $d_{ij}$ where $i\in(1,m)$ and $j\in(1,n)$. Conventionally, each job has a predetermined processing order that has to be followed to complete all operations. The order of a particular job can be represented as: 
\vspace{0.25cm}
\begin{equation}\label{eq:jobRepresentation}
    J_j=\{O_{j1}d_{j1}, O_{j2}d_{j2}, \dots, O_{jm}d_{jm}\} \text{, for } j \in (1,n), m\in(1,m).\
\vspace{0.25cm}
\end{equation}

Based on the problem definition, the machining sequence is developed and the quality of the solution is evaluated by the makespan value. The machine up-time can be calculated using $T_i = \sum_{j=1}^{n}d_{ij}$  and the free time or idling time by  $F_i= \sum_{j=1}^{n}f_{ij}$ , which correlates with the makespan of the solution $C_{\mathrm{max}} = \underset{i}{\max}{(\sum_{j=1}^{n}d_{ij} + f_{ij})}$. The difficulty in finding the global optimum solution (lower bound) increases exponentially with the problem size $n \times m$.

\subsection{\acf{ppo}}
\ac{ppo} is a policy gradient method that uses sampling data obtained from environment interaction in order to optimize the surrogate objective function using stochastic gradient ascent \citep{schulman2017proximal}. Unlike Q-learning or \ac{trpo}, the \ac{ppo} algorithm is more data efficient, robust and less complex to implement. The surrogate objective $L^\mathrm{\emph{CLIP}}$ of the \ac{ppo} is given by
\vspace{0.25cm}
\begin{equation}\label{eq:surrogateObjective}
    L^\mathrm{\emph{CLIP}} (\theta) = \widehat{E}_t \left[ \min{⁡(r_t(\theta)\widehat{A}_t, \mathrm{\emph{clip}}(r_t(\theta),1-\epsilon,1+\epsilon)}\widehat{A}_t) \right ].
\end{equation}

\ac{ppo} uses a modified surrogate objective function that retrieves the pessimistic bound of the unclipped objective. Finding the pessimistic bound is done by comparing, and finding the minimum among the unclipped objective $r_t(\theta) \widehat{A}_t$  and clipped objective $\mathrm{\emph{clip}}(r_t (\theta),1-\epsilon,1+\epsilon) \widehat{A}_t$. By modulating the hyperparameter $\epsilon$, the range of the update can be limited.

\section{Related Works}
Although research adressing \ac{jssp}s is rather sparse, several different algorithms have been employed to attain their specific optimization goals. Algorithms such as Taboo search \citep{taillard1994parallel}, simulated Annealing \citep{van1992job}, genetic algorithms and particle swarm optimization \citep{pezzella2008genetic} have proven to solve the problem, but lack in either computation time or generalization capabilities.
Advancements in \ac{drl} approaches in recent years have enabled considerable progress for the domain of \ac{cop} applications \citep{cappart2021combining, oren2021solo}. Some of the major \ac{cop}s have been successfully solved using \ac{drl} such as the \ac{tsp} \citep{zhang2021solving, d2020learning, zhang2020deep}, the Knap Sack Problem \citep{afshar2020state, cappart2021combining} and the Steiner Tree Problem \citep{du2021vulcan}. \citet{zhang1995reinforcement} were able to show the potential of \ac{rl} for \ac{jssp}s as far back as 1995, by improving the results of the scheduling algorithm by \citet{deale1994space} which used a temporal difference algorithm in combination with simulated annealing. Further, the study from \citet{gabel2012distributed} on using a gradient descent policy search method for scheduling problems demonstrated the feasibility of \ac{drl} in \ac{jssp}s. Despite the reduced computation time, the solution found was not better than that of traditional solvers. This limitation was partially overcome by \citet{liu2020actor}, who designed an environment based on a \ac{mmdp} and used a \ac{ddpg} for their approach. The agent performed well on the smaller instances, producing a good scheduling score of around 90\% but eventually, the performance declined with the increase in size of the instances. To deal with the increased complexity of the problem, an Adaptive Job Shop Scheduling based on a Dueling Double Deep Q-Network with a prioritized reply was proposed by \citet{han2020research}. The authors used a disjunctive graph-based model to design the environment and transform it into a sequential decision-making problem. The algorithm was tested for generalization ability by training the network with random events in the existing environment to quickly adapt to new problems. However, the agent was not tested with a completely new dataset, an issue which was overcome by \citet{zhang2020learning}. The authors developed a graph neural network which enabled them to solve size-agnostic problems. Similar to the previous approach the authors used a disjunctive $graph = (\partial,C,D)$ with $\partial$ being the chosen operation of the \ac{jssp} $\partial=\{O_{ij}| \forall_{i,j}\}\cup\{S,T\}$, $C$ the set of directed arcs and $D$ the set of undirected arcs, to represent the state space of the \ac{jssp}. The performance of the agent was promising, although the generalized results were far off from the optimum. To handle this, \citet{tassel2021reinforcement} proposed a new \ac{drl} algorithm to solve the \ac{jssp}s with compact state space representation and a simple dense reward function. The action space was designed for $n$ jobs along with an additional job called No-Op (No Operation).  The agent was tested with different benchmark instances where the agent performed around 18\% better than  \citet{zhang2020learning} and 10\% better than \citet{han2020research}. Even though they provide a near-optimum solution, the approach falls short of the generalization objective. The motive of our approach is to develop an efficient environment that can perform better in both the objectives i.e. generalization and near-optimum solutions.

\section{Methodologies}
The job shop environment is built with \emph{OpenAI} gym which provides modules to develop a reinforcement learning environment. An agent learns to solve the environment and optimize the parameters of the policy by interacting with it through actions.
\subsection{Environment Outline}
\begin{figure}[h]    
    \centering
    \def\svgwidth{\textwidth}
    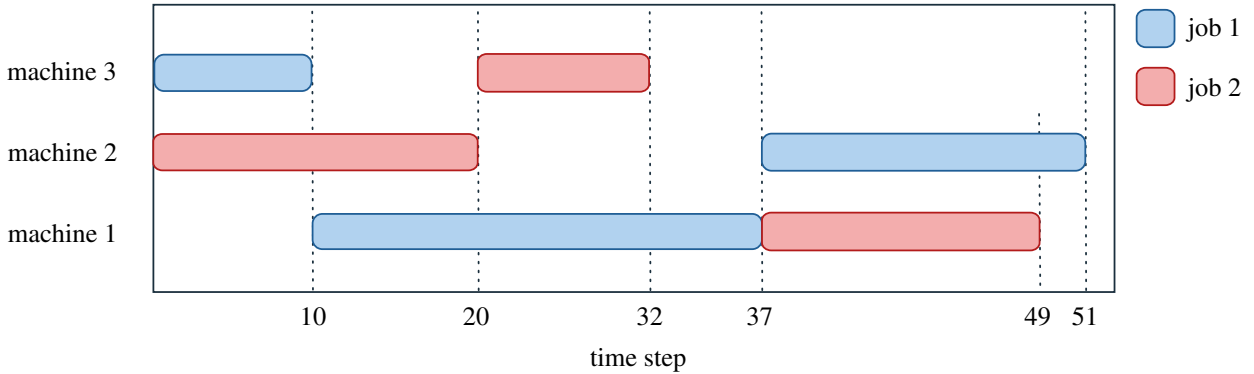 
    \caption{Job sequence on machines for a job shop problem with 3 machines and 2 jobs)}
    \label{fig:jobSequence}
\end{figure}

Along with the mathematical constraints explained before, the environment is designed with further constraints \citep{blazewicz2001scheduling, pinedo2012scheduling}:
\begin{enumerate}
    \item No pre-emption is allowed i.e. operations cannot be interrupted
    \item Each machine can  handle only one job at a time
    \item No precedence constraints among operations of different jobs
    \item Fixed machine sequences of each job
\end{enumerate}
For example, when considering a \ac{jssp} with three machines and two jobs with a job order  $J_1=\{3, 1, 2\}, J_2=\{2, 3, 1\}$ the environment treats each operation as an atomic operation (cannot be interrupted). So once the job $J_0$ is assigned on machine $M_3$ at time step $0$ it cannot be interrupted till time step $10$, as shown in \emph{Figure~\ref{fig:jobSequence}}. 
Based on the described problem definition, we consider the \ac{jssp} as a single agent problem. Additionally, it has been shown by \citet{tassel2021reinforcement} that the performance of the single agent \ac{drl} outperforms existing state-of-the-art \ac{drl} methods. Unlike Tassel et al.'s approach, our implementation of the environment has a different environmental design with no additional action (No-Op) and no non-final prioritization technique.
\subsection{Time Step Transition}
For the efficient learning of the agent, the time step and machines to be assigned are chosen based on the eligibility of operation. This enables the environment to provide the agent with only the potential requests at which machines can be assigned. At each chosen time step, the environment fetches all eligible operations $O$ based on the provided job order of the problem. The definition of an eligible operation relies on the job order and the current state of the jobs and machines.
For example, consider the previously discussed job shop problem with three machines and two jobs, with total operations $m \times n = 6$. The operations can be expressed as $O={O_{11} \rightarrow J_1 M_3,O_{12} \rightarrow J_1 M_1,O_{13} \rightarrow J_1 M_2,O_{21} \rightarrow J_2 M_2,O_{22} \rightarrow J_2 M_3,O_{23} \rightarrow J_2 M_1}$ along with the processing time $d={d_{11} \rightarrow 10,d_{12} \rightarrow 27,d_{13} \rightarrow 14, d_{21} \rightarrow 20, d_{22} \rightarrow 12, d_{23} \rightarrow 12}$. Based on the predetermined job order, the eligible operations are $O_{11}$ and $O_{21}$ at time step $0$ and similarly, at time step $10$, the eligible operation is limited to $O_{12}$. Likewise, the time step where the agent will be queried next in the environment is chosen based on the minimum length of the operation which is currently active. This enables the program to skip unnecessary checks for run time length in order to determine the next operation. At time step $0$, the active operations are $O_{11}$, $O_{21}$ with corresponding times $d_{11}$, $d_{21}$ of length $10$ and $20$. The environment then directly jumps to time step $10$ which is minimum operation length of the currently running operations $O_{11}$, $O_{21}$. 
The transition is further regulated by the availability of the machines and jobs at the future time step. For instance, at time step $20$, operations $O_{12}$, $O_{22}$ are active until time step $32$ and $37$. Even though the operation $O_{22}$ has the minimum processing time, jumping to time step $32$ will not be useful, since the next operation $O_{23}$ which involves processing job $J_2$ on machine $M_1$ is not eligible at time step $32$ as the machine $M_1$ is processing job $J_1$ till $37$. So the environment directly jumps to the time step $37$. 
With this mode of time step transition and querying mechanism, the agent was able to solve the \ac{jssp} environment by taking steps approximately equal to the total number of operations. To put this into perspective, the \citet{taillard1993benchmarks} ta01 instance with $15$ machines and $15$ jobs was solved by the agent with $225$ requests which is equivalent to the total number of operations. Exploiting this mechanism, a significant improvement can be reached.
\subsection{Action Space}
The environment is controlled by a single discrete action space. With this action space, the agent determines the suitable job to process for a particular machine at each step. The agent is constrained to the set of jobs available $A_t=\{J_1, J_2, J_3, \dots, J_n\}$. 

\subsection{States}
At each time step, the state space is updated with the current status information of all jobs and machines. We developed a dictionary state space which incorporates the following information: (1) status of the machines – a boolean vector of size $m$ indicating which machines are busy and idling, (2) operation progression – a vector of size $m$ holding information on whether or not an operation is still running, (3) current remaining jobs – a vector of size $n$ that shows the remaining operations per job, (4) overall operation overview – a two dimensional boolean array of size $n\times m$ that provides the status of the operations, (5) availability of jobs – a vector of size $n$ that indicates the next eligible operation, (6) current machine processing information – a vector of size $m$ that holds information regarding the currently processed jobs on the machine. 

\subsection{Reward}
The reward function must closely correspond to the scheduling goal, e.g. guiding the suitable assignment of jobs to the appropriate machines and reducing the makespan of the schedule. It has been clearly shown in several studies that the performance of an agent with dense rewards is better than the performance of an agent with sparse rewards \citep{mohtasib2021study}. In this work, we have designed a simple dense reward function $R$ to provide a feedback regarding the operation assignment and a final reward $R_f$ to express the viability of the achieved goal. Therefore the cumulative reward equals

\begin{equation}
    R_{\mathrm{cumulative}} = \sum_a{R+R_f}
\end{equation}

\begin{equation}
    R(a_t, s_t) = \left\{
	                   \begin{array}{lll}
		                  1, O & \in & \partial \mathrm{\emph{ valid assignment}} \\
		                  0, O & \notin & \partial \mathrm{\emph{ invalid assignment}}
	                   \end{array}
                \right.
\end{equation}

The final reward $R_f$ can only be achieved if the agent is successful in assigning all operations. The $R_f$ is closely related to the makespan that is obtained using the policy of the agent.
\vspace{0.25cm}
\begin{equation}
    R_f = (\mathrm{\emph{roll out}} + 100 - C_{max}^*) \cdot C_{0}.
\vspace{0.25cm}
\end{equation}
The constants in $R_f$ are simply a scaling factor that help the agent to improve its problem solving performance and are found by hyperparameter optimization .

\subsection{Markov Decision Process Formulation}
The \ac{jssp} can be modelled as a \ac{mdp} since the assignment of the job sequentially changes the environment in terms of states and rewards, therefore the Markov property is fulfilled. \ac{jssp}s have been formulated as \ac{mdp}s in several previous studies \citep{singh1997dynamically, zhang2017real, zhang2020learning, tassel2021reinforcement, han2020research} with different approaches based on the type of algorithm used to solve the problem. The agent assigns a job through an action $a_t$ at time step $t$ and retrieves the next state $s_{t+1}$. Unlike single assignment at a time step $t$, there can also be multiple assignment based on the number of eligible operations $k$ at time step $t$. The actions at time step $t$ are given by $a_{t_0}, a_{t_1}, \dots, a_{t_k}$ extending the state space at $t$ by $s_{t_0}, s_{t_1}, \dots, s_{t_k}$. The probability of the next state is modified based on the sub actions taken at the time step. This can be expressed by

\vspace{0.25cm}
\begin{equation}
    p(s',r|s,a) = Pr\{S_t = s' , R_t = r|S_{t-1} = s, A_{t-1} = a\}. 
\end{equation}
\vspace{0.25cm}

\subsection{Generalization}
\begin{figure}
        \centering
            \def\svgwidth{\textwidth}
                    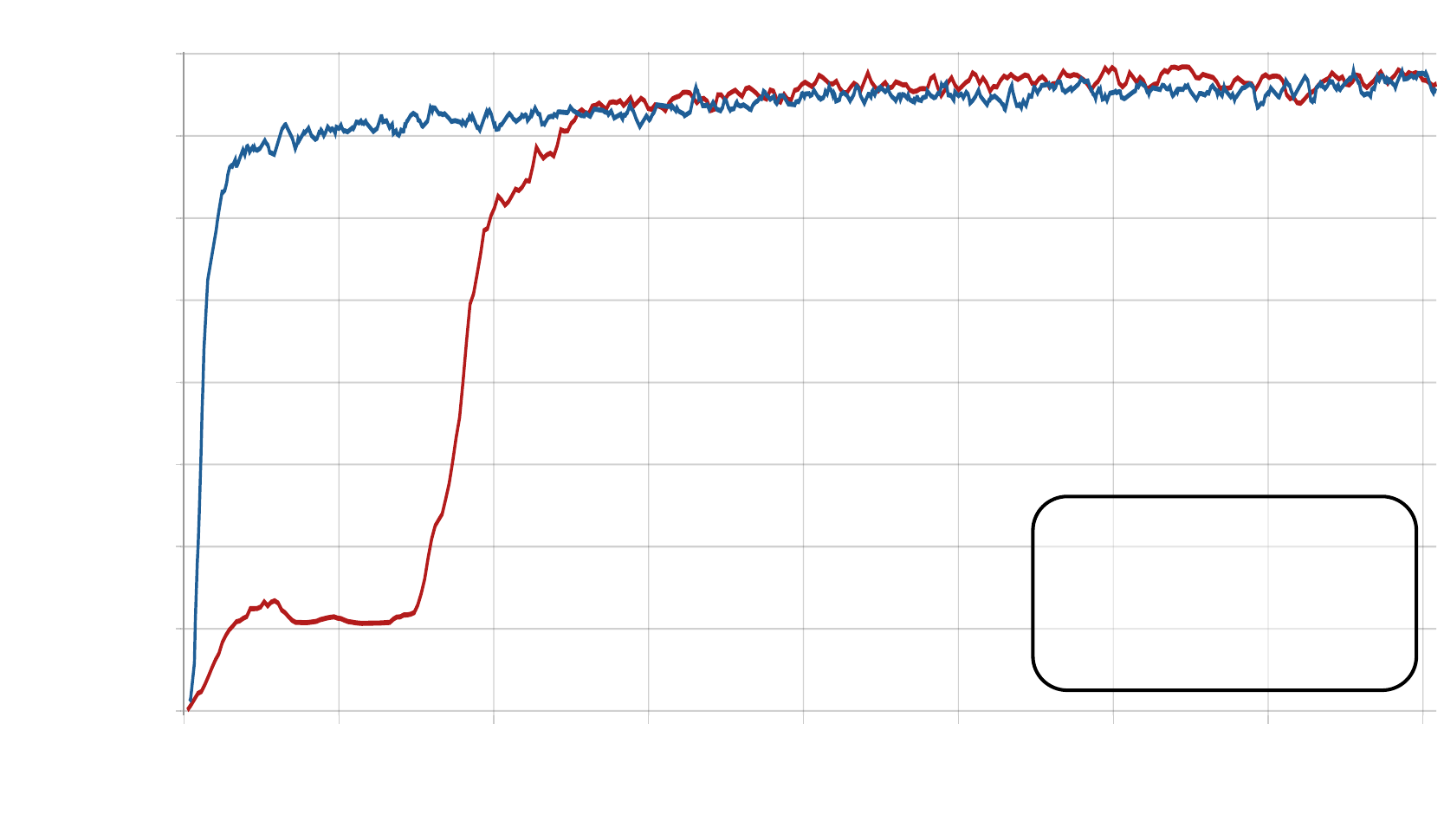
    \caption{Comparison between \ac{osm} implementation (red) and conventional environment (blue) - performance  of the agent vs training steps}
    \label{fig:comparison_left}
    \hfill
    \centering
        \def\svgwidth{\textwidth}
                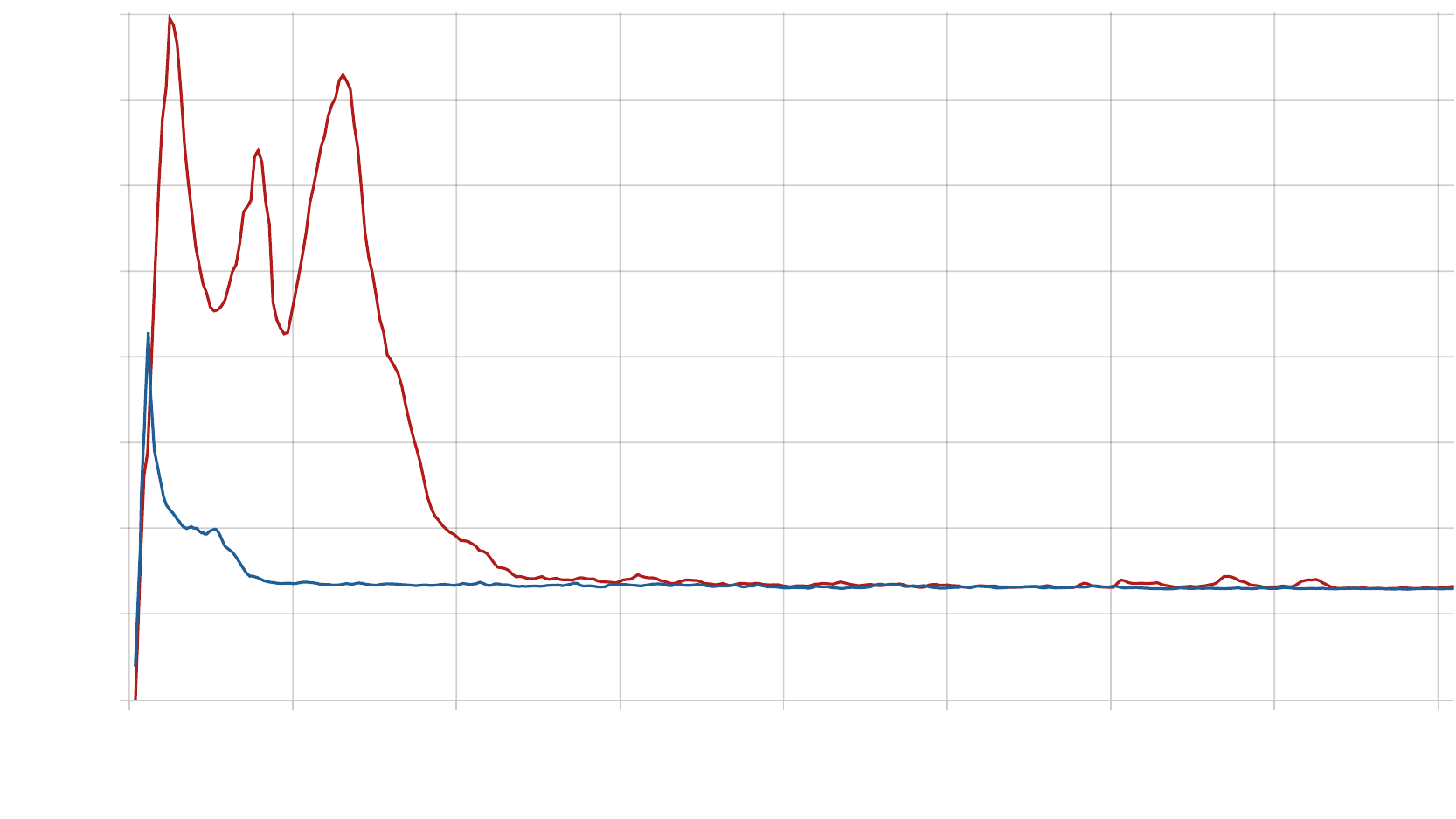
    \caption{Comparison between \ac{osm} implementation (red) and conventional environment (blue) - episode length vs training steps}
    \label{fig:comparison_right}
\end{figure}

In order to increase the agents generalization capability, we introduce an \ac{osm} to our environment. The agent is trained with a particular instance along with an \ac{osm} and then evaluated with another instance of the same size. We used ta01 benchmark \citep{taillard1993benchmarks} with $15\times 15$ problem size leading to:
\begin{equation}
    \text{\# of swaps} = T_p \cdot \frac{m \times n}{100} \cdot \tau .   
\end{equation}\label{eq:swaps}
 \newline
\noindent
The objective of the method is to swap the processing order exponentially during the training phase $T_p$ of the agent. The training phase $T_p$ is dissected based on the episodic termination i.e. for the $15 \times 15$ job shop instance, initially the agent might take $1000$ time steps to reach the episodic termination and so here $T_p$  will increment by $1$ but at the end of the training, the agent will reach episodic termination within $225$ steps which will also increment $T_p$ by only $1$. Thus the episodic length and the training steps cannot be linearly correlated. This phenomena can clearly be observed in \emph{Figure~\ref{fig:comparison_right}} in which the episode length decays with increasing training steps. In the conventional environment (no \ac{osm}), the agent was quickly able to converge as the environment is static. In the \ac{osm} case, the environment behaves somewhat randomly initially and converges as the training phase progresses. 
The generalization performance is dependent on the amount of randomness provided during the training. A very high \ac{osm} level in the training caused the agent to fail to learn the primary objective of scheduling. On the other hand a very low \ac{osm} level resulted in high validation loss. The execution rate  parameter $\tau$ along with $T_p$ constrains the \ac{osm} execution over the period of training steps (\ref{eq:swaps}). We set $\tau = 0.00667$ which implicitly provides approximately 10\% \ac{osm} in the 15 $\times$ 15 problem running over 6 million steps.
\begin{figure}
      \centering
            \def\svgwidth{\textwidth}
                    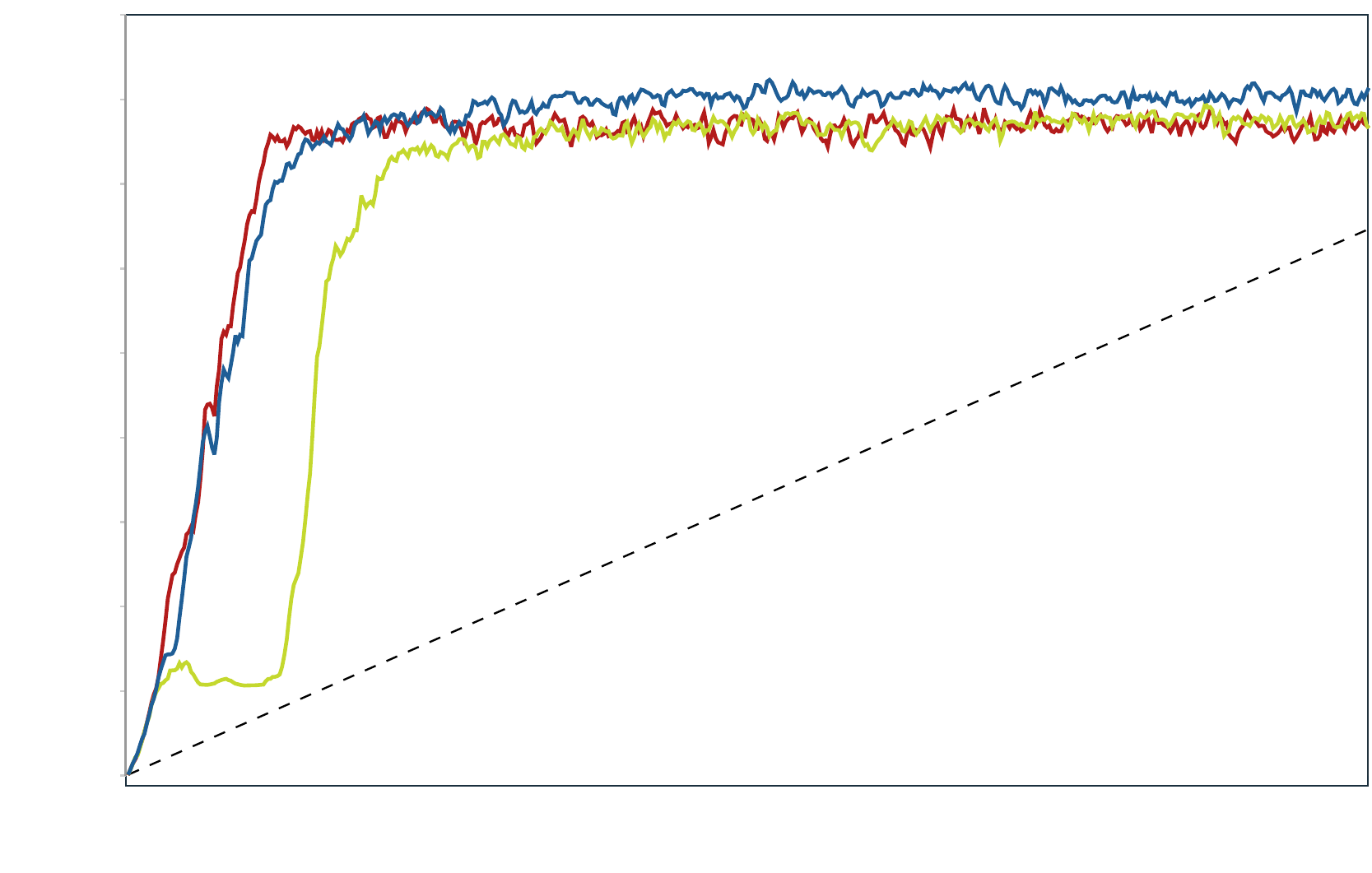
         \caption{Performance with different execution rates - agent performance vs time steps $(\text{blue} : \tau = 0.01, \text{green} : \tau = 0.00667, \text{red} : \tau = 0.005, \text{black} - T_p)$}
        \label{fig:performance_left}
        \hfill
        \def\svgwidth{\textwidth}
                    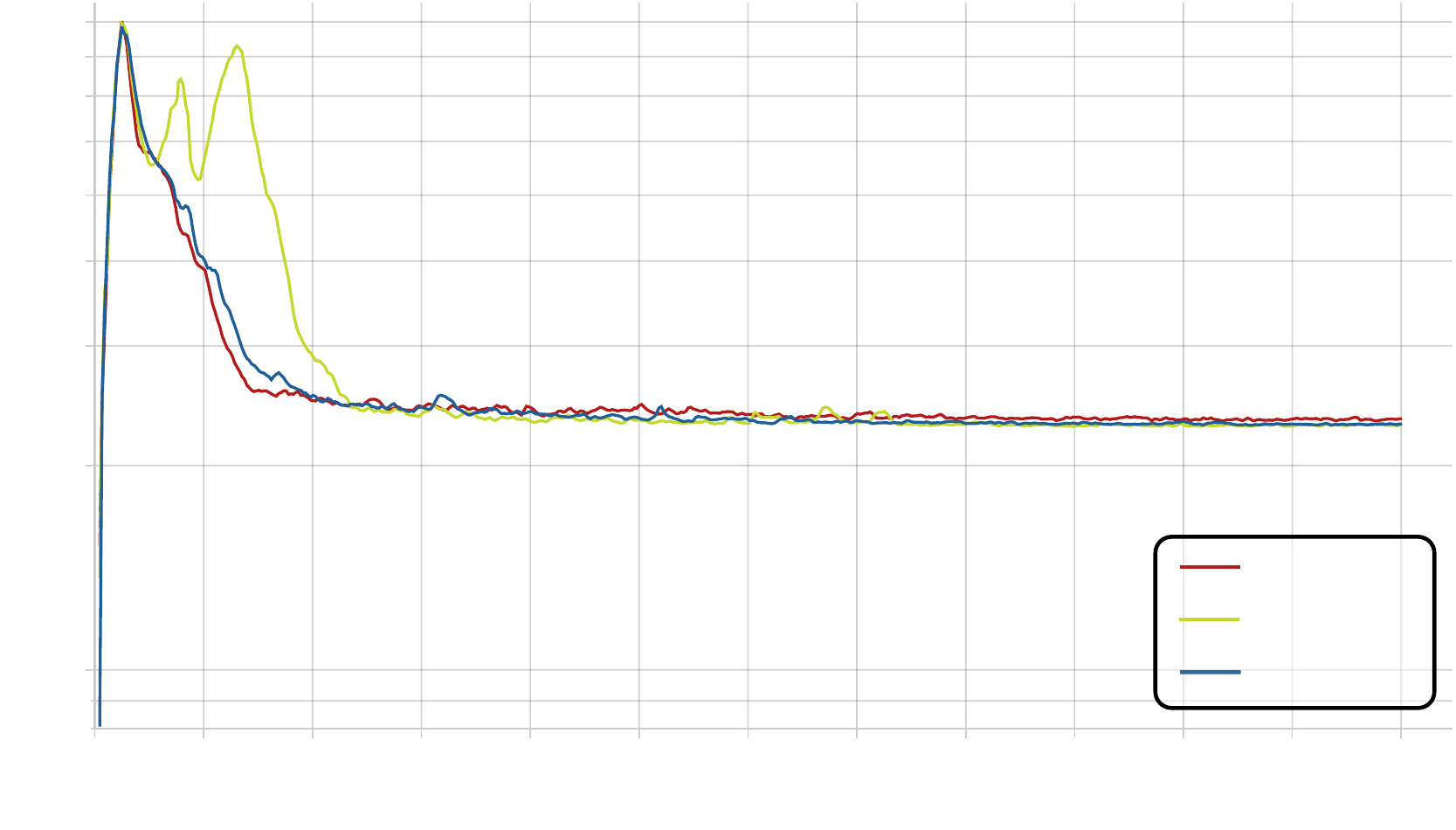
         \caption{Performance with different execution rates - episode length vs training steps $(\text{blue} : \tau = 0.01, \text{green} : \tau = 0.00667, \text{red} : \tau = 0.005$}
        \label{fig:performance_right}
\end{figure}
The performance of the agent during the training with the environment of $\tau=0.00667$ is lower than the $\tau=0.005$ (5\% change to the true instance) but in contrast to that, the performance on the test dataset is better for the $\tau=0.00667$ (10\% change to the true instance). This indicates that the agent is more generalized when trained with 10\% randomness provided during the training. The parameter is tremendously sensitive in a way that increasing the randomness by more than 15\% lead to a non-converging training progress collapse.
\section{Experiments}
Our proposed strategy was implemented using OpenAI Gym toolkit \citep{brockman2016openai} and Stable-Baselines3 \citep{raffin2021stable} which provide \ac{rl} environment APIs and reliable reinforcement learning algorithms.

\subsection{Model Configuration}
The policy network is designed as an actor network with two hidden layers and a value network that also has two hidden layers, both of size $256$. The hyperparameter optimization was carried out using the optuna optimizer \citep{akiba2019optuna}. We set the clipping parameter to $0.2$ and the discount factor $\gamma$ to  $966 \times 10^{-3}$. In order to avoid major updates in the network at the end, we introduced a linear scheduler for the learning rate which decays from $1 \times 10^{-4}$ to $1 \times 10^{-8}$. The policy update step is set based on the size of the problem. This parameter is sensitive towards the number of steps taken to solve the environment which is dependent on the size of the problem. For example, in the $15 \times 15$ instance, it was set to $448$. Finally, we developed a roll out parameter for the environment which is also dependent on the problem size which helps to reduce the training time. The roll out parameter indicates when to terminate the current training phase.

\subsection{Training}
The training process was carried out in two modes, one for each instance and generalized instance size. The efficiency of the solution is analyzed by using the previously obtained upper bounds for these instances. Additionally, to analyze the performance of the agent, the environment provides an occupancy cumulative average value for the job assignments. Through this, we were also able to speed up the training process by setting an occupancy threshold value which needs to be satisfied by the agent to achieve its goal, if it doesn't, the environment terminates. By implementing this technique, the network was able to converge quicker and required less training time to reach an efficient solution.

\subsection{Benchmark Instances}
To evaluate the performance, we have used the commonly used benchmark instances in this field of study. \emph{Table~\ref{tab:perfComp}} presents an overview of the instances that were used for training and evaluation. We mainly compared our performance using \citet{taillard1993benchmarks} and \citet{demirkol1998benchmarks} instances with \citet{han2020research}, \citet{zhang2020learning} and \citet{tassel2021reinforcement}, as they used same benchmark instances. For our generalized approach, we compare our results with \citet{zhang2020learning}, even though we are not size agnostic, we achieved partial generalization in terms of problem size.
\begin{table}[]
\centering
\begin{tabular}{|c|c|}
\hline
\textbf{Authors}                    & \textbf{Instance size}   \\ \hline
\citet*{adams1988shifting}    & 10 $\times$ 10, 20 $\times$ 15         \\ \hline
\citet*{demirkol1998benchmarks} & 20 $\times$ 15 to 50 $\times$ 20        \\ \hline
\citet*{fisher1963probabilistic}       & 6 $\times$ 6, 10 $\times$ 10,   20 $\times$ 5 \\ \hline
\citet*{lawrence1984resouce}                  & 10 $\times$ 10 to 15 $\times$ 15        \\ \hline
\citet*{applegate1991computational}      & 10 $\times$ 10                  \\ \hline
\citet*{taillard1993benchmarks}                  & 15 $\times$ 15 to 20 $\times$ 100    \\ \hline
\citet*{yamada1992genetic}          & 20 $\times$ 20                  \\ \hline
\citet*{storer1992new}     & 20 $\times$ 10 to 50 $\times$ 10        \\ \hline
\end{tabular}
\vspace{0.2cm}
\caption{Benchmark instances}
\label{tab:benchInst}
\end{table}
\subsection{Results}
We compare our results with the existing state-of-the-art algorithms and with common heuristics. \emph{Table~\ref{tab:perfComp}} provides an overview of the corresponding performances. The agent was able to perform better for small size instances and achieved comparable performance for large scale instances. Even though the agent was not able to perform better than the state-of-the-art \ac{drl} approach by \citet{tassel2021reinforcement}, the goal of this study was to develop a generalized agent which could achieve a good performance without having trained on an instance of the same size.

\begin{table}[h!]
\centering
\begin{tabular}{lllllllll}
\multirow{2}{*}{\textbf{Instance}} &
  \textbf{Size} &
  \multirow{2}{*}{\textbf{MWKR}} &
  \multirow{2}{*}{\textbf{SPT}} &
  \multirow{2}{*}{\textbf{\citeauthor{tassel2021reinforcement}}} &
  \multirow{2}{*}{\textbf{\citeauthor{han2020research}}} &
  \multirow{2}{*}{\textbf{\citeauthor{zhang2020learning}}} &
  \multirow{2}{*}{\textbf{Ours*}} &
  \multirow{2}{*}{\textbf{Lower Bound}} \\
      & \textbf{(n $\times$ m)} &      &      &      &      &      &      &      \\
Ft06  & 6 $\times$ 6            & -    & -    & -    & -    & -    & 55*  & 55   \\
La05  & 10 $\times$ 5           & 787  & 827  & -    & 593  & -    & 593* & 593  \\
La10  & 15 $\times$ 5           & 1136 & 1345 & -    & 958  & -    & 958* & 958  \\
La16  & 10 $\times$ 10          & 1238 & 1588 & -    & 980  & -    & 974  & 945  \\
Ta01  & 15 $\times$ 15          & 1786 & 1872 & -    & 1315 & 1443 & 1352 & 1231 \\
Ta02  & 15 $\times$ 15          & 1944 & 1709 & -    & 1336 & 1544 & 1354 & 1244 \\
dmu16 & 30 $\times$ 20          & 5837 & 6241 & 4188 & 4414 & 4953 & 4632 & 3751 \\
dmu17 & 30 $\times$ 20          & 6610 & 6487 & 4274 & -    & 5579 & 5104 & 3814 \\
Ta41  & 30 $\times$ 20          & 2632 & 3067 & 2208 & 2450 & 2667 & 2583 & 2005 \\
Ta42  & 30 $\times$ 20          & 2401 & 3640 & 2168 & 2351 & 2664 & 2457 & 1937 \\
Ta43  & 30 $\times$ 20          & 3162 & 2843 & 2086 & -    & 2431    & 2422 & 1846
\end{tabular}
\vspace{0.2cm}
\caption{Performance comparison of the conventional env model, * indicates the solution is optimal}
\label{tab:perfComp}
\end{table}
\subsection{Generalized Result}
Based on the generalization research, \citet{zhang2020learning}'s approach using graph neural networks has produced promising results. The training time for larger instances is drastically reduced using their size-agnostic network. Although we are not size agnostic, we developed our generalization approach with respect to the problem size. Through this approach, we were able to produce better results with reduced execution time since the training is necessary only once with a particular problem size. We have compared our results based on different problem sizes with \citet{taillard1993benchmarks} and \citet{demirkol1998benchmarks}. We tried three different execution rates $\tau = 0.01$, $\tau = 0.00667$,  $\tau = 0.005$ which impose 15\%, 10\%, 5\% swaps in the original dataset. 
\begin{table}[h!]
\centering
\begin{tabular}{lllllllll}
\multirow{2}{*}{\textbf{Instance}} &
  \textbf{Ta01-OSM} &
  \textbf{Ta01-OSM} &
  \textbf{Ta01-OSM} &
  \multirow{2}{*}{\textbf{MWKR}} &
  \multirow{2}{*}{\textbf{SPT}} &
  \multirow{2}{*}{\textbf{Ours*}} &
  \multirow{2}{*}{\textbf{\citeauthor{zhang2020learning}}} &
  \multirow{2}{*}{\textbf{Lower Bound}} \\
     & \textbf{With 5\%} & \textbf{With 10\%} & \textbf{With 15\%} &      &      &               &      &      \\
Ta02 & 1491              & \textbf{1486}      & 1546               & 1944 & 1709 & \textbf{1354} & 1544 & 1244 \\
Ta03 & 1443              & \textbf{1437}      & 1525               & 1947 & 2009 & \textbf{1388} & 1440 & 1218 \\
Ta04 & 1568              & \textbf{1502}      & 1614               & 1694 & 1825 & \textbf{1513} & 1637 & 1175 \\
Ta05 & 1599              & \textbf{1481}      & 1483               & 1892 & 2044 & \textbf{1443} & 1619 & 1224 \\
Ta06 & 1776              & \textbf{1507}      & 1552               & 1976 & 1771 & \textbf{1360} & 1601 & 1238 \\
Ta07 & 1526              & \textbf{1500}      & 1605               & 1961 & 2016 & \textbf{1354} & 1568 & 1227 \\
Ta08 & 1631              & \textbf{1540}      & 1524               & 1803 & 1654 & \textbf{1377} & 1468 & 1217 \\
Ta09 & 1662              & \textbf{1664}      & 1597               & 2215 & 1962 & \textbf{1401} & 1627 & 1274 \\
Ta10 & 1573              & \textbf{1524}      & 1659               & 2057 & 2164 & \textbf{1370} & 1527 & 1241
\end{tabular}
\vspace{0.2cm}
\caption{Generalized Taillard's $15 \times 15$ instance results with various OSM execution rate}
\label{tab:genTaill15x15}
\end{table}
For the problem size $15 \times 15$, the agent was trained with Taillard’s 01 and tested with other $15 \times 15$ instances of Taillard's. It can be clearly observed from \emph{Table~\ref{tab:genTaill15x15}} that, with constrained randomness, the agent was able to generalize better and produce near optimum solutions.
\begin{table}[h!]
\centering
\begin{tabular}{lllllllll}
\multirow{2}{*}{\textbf{Instance}} &
  \textbf{Ta41-OSM} &
  \textbf{Ta41-OSM} &
  \textbf{Ta41-OSM} &
  \multirow{2}{*}{\textbf{MWKR}} &
  \multirow{2}{*}{\textbf{SPT}} &
  \multirow{2}{*}{\textbf{Ours*}} &
  \multirow{2}{*}{\textbf{\citeauthor{zhang2020learning}}} &
  \multirow{2}{*}{\textbf{Lower Bound}} \\
     & \textbf{With 5\%} & \textbf{With 7.5\%} & \textbf{With 10\%} &      &      &               &      &      \\
Ta42 & 2903              & 2831                & \textbf{2572}      & 3394 & 3640 & \textbf{2457} & 2664 & 1937 \\
Ta43 & 2800              & 2651                & \textbf{2614}      & 3162 & 2843 & \textbf{2422} & 2431 & 1846 \\
Ta44 & 2991              & 2751                & \textbf{2745}      & 3388 & 3281 & \textbf{2598} & 2714 & 1979 \\
Ta45 & 2851              & 2812                & \textbf{2692}      & 3390 & 3238 & \textbf{2587} & 2637 & 2000 \\
Ta46 & 2986              & 2842                & \textbf{2674}      & 3268 & 3352 & \textbf{2606} & 2776 & 2006 \\
Ta47 & 2854              & 2807                & \textbf{2677}      & 2986 & 3197 & \textbf{2538} & 2476 & 1889 \\
Ta48 & 2758              & 2753                & \textbf{2638}      & 3050 & 3445 & \textbf{2461} & 2490 & 1937 \\
Ta49 & 2800              & 2646                & \textbf{2566}      & 3172 & 3201 & \textbf{2501} & 2556 & 1961 \\
Ta50 & 2887              & 2654                & \textbf{2616}      & 2978 & 3083 & \textbf{2550} & 2628 & 1923
\end{tabular}
\vspace{0.2cm}
\caption{Generalized Taillard's $30 \times 20$ instance results with various OSM}
\label{tab:genTaill30x20}
\end{table}
\begin{table}[h!]
\centering
\begin{tabular}{lllllllll}
\multirow{2}{*}{\textbf{Instance}} &
  \textbf{Ta41-OSM} &
  \textbf{Ta41-OSM} &
  \textbf{Ta41-OSM} &
  \multirow{2}{*}{\textbf{MWKR}} &
  \multirow{2}{*}{\textbf{SPT}} &
  \multirow{2}{*}{\textbf{Ours*}} &
  \multirow{2}{*}{\textbf{\citeauthor{zhang2020learning}}} &
  \multirow{2}{*}{\textbf{Lower Bound}} \\
      & \textbf{With 5\%} & \textbf{With 7.5\%} & \textbf{With 10\%} &      &      &               &      &      \\
Dmu16 & 5413              & 5560                & \textbf{4907}      & 5837 & 6241 & \textbf{4632} & 4953 & 3751 \\
Dmu17 & 5926              & 5911                & \textbf{5646}      & 6610 & 6487 & \textbf{5104} & 5379 & 3814 \\
Dmu18 & 5380              & 5773                & \textbf{5287}      & 6363 & 6978 & \textbf{4998} & 5100 & 3844 \\
Dmu19 & 5236              & 5136                & \textbf{4993}      & 6385 & 5767 & \textbf{4759} & 4889 & 3768 \\
Dmu20 & 5263              & 5318                & \textbf{5131}      & 6472 & 6910 & \textbf{4697} & 4859 & 3710
\end{tabular}
\vspace{0.2cm}
\caption{Generalized Demikrol $30 \times 20$ instance results with various OSM}
\label{tab:genDemikrol30x20}
\end{table}
The agent's performance in a conventional environment is better than its performance in a \ac{osm}- environment. However the training time required to train 10 instances of the same size is reduced by a factor of 10 when using the \ac{osm}-environment. This is due to the fact that the training progress of the first instance can be transferred to the remaining nine instances.

\section{Conclusion}
In this paper, we have developed a reinforcement learning environment that solves \ac{jssp}s. Moreover we showed the increased generalization capability when employing our \ac{osm} implementation. The main motive of our work was to develop a generalized model that can provide near optimum solutions. Even though, our approach is not fully generalized like \citet{zhang2020learning} we provide a size dependent generalization which is of high relevance for the industry. Based on the generalized result, it is clear that our approach outperforms the \ac{pdr} based \ac{drl} approach by \citet{zhang2020learning}. Based on single instance training, our results show that our agent performed similarly to the state of the art \ac{drl} algorithms. For our future work, we plan to modify this approach to be size-agnostic which can be used by the industry to obtain more reliable scheduling results.     

\section*{Acknowlegements}
This project is supported by the Federal Ministry for Economic Affairs and Climate Action (BMWK) on the basis of a decision by the German Bundestag.

\newpage

\bibliographystyle{unsrtnat}
\bibliography{references}  
\end{document}

%% file: figures/schedule.pdf_tex
\begingroup%
  \makeatletter%
  \providecommand\color[2][]{%
    \errmessage{(Inkscape) Color is used for the text in Inkscape, but the package 'color.sty' is not loaded}%
    \renewcommand\color[2][]{}%
  }%
  \providecommand\transparent[1]{%
    \errmessage{(Inkscape) Transparency is used (non-zero) for the text in Inkscape, but the package 'transparent.sty' is not loaded}%
    \renewcommand\transparent[1]{}%
  }%
  \providecommand\rotatebox[2]{#2}%
  \newcommand*\fsize{\dimexpr\f@size pt\relax}%
  \newcommand*\lineheight[1]{\fontsize{\fsize}{#1\fsize}\selectfont}%
  \ifx\svgwidth\undefined%
    \setlength{\unitlength}{493.23271504bp}%
    \ifx\svgscale\undefined%
      \relax%
    \else%
      \setlength{\unitlength}{\unitlength * \real{\svgscale}}%
    \fi%
  \else%
    \setlength{\unitlength}{\svgwidth}%
  \fi%
  \global\let\svgwidth\undefined%
  \global\let\svgscale\undefined%
  \makeatother%
  \begin{picture}(1,0.29660403)%
    \lineheight{1}%
    \setlength\tabcolsep{0pt}%
    \put(0,0){\includegraphics[width=\unitlength,page=1]{./figures/schedule.pdf}}%
    \put(-1.49561723,-0.5412007){\color[rgb]{0.34901961,0.34901961,0.34901961}\makebox(0,0)[lt]{\lineheight{1.25}\smash{\begin{tabular}[t]{l}0\end{tabular}}}}%
    \put(-1.51572355,-0.49960098){\color[rgb]{0.34901961,0.34901961,0.34901961}\makebox(0,0)[lt]{\lineheight{1.25}\smash{\begin{tabular}[t]{l}500\end{tabular}}}}%
    \put(-1.5257719,-0.45800126){\color[rgb]{0.34901961,0.34901961,0.34901961}\makebox(0,0)[lt]{\lineheight{1.25}\smash{\begin{tabular}[t]{l}1000\end{tabular}}}}%
    \put(-1.5257719,-0.41595899){\color[rgb]{0.34901961,0.34901961,0.34901961}\makebox(0,0)[lt]{\lineheight{1.25}\smash{\begin{tabular}[t]{l}1500\end{tabular}}}}%
    \put(-1.5257719,-0.37435927){\color[rgb]{0.34901961,0.34901961,0.34901961}\makebox(0,0)[lt]{\lineheight{1.25}\smash{\begin{tabular}[t]{l}2000\end{tabular}}}}%
    \put(-1.5257719,-0.33275955){\color[rgb]{0.34901961,0.34901961,0.34901961}\makebox(0,0)[lt]{\lineheight{1.25}\smash{\begin{tabular}[t]{l}2500\end{tabular}}}}%
    \put(-1.48679065,-0.56509841){\color[rgb]{0.34901961,0.34901961,0.34901961}\makebox(0,0)[lt]{\lineheight{1.25}\smash{\begin{tabular}[t]{l}Ta01\end{tabular}}}}%
    \put(-1.4334127,-0.56509841){\color[rgb]{0.34901961,0.34901961,0.34901961}\makebox(0,0)[lt]{\lineheight{1.25}\smash{\begin{tabular}[t]{l}Ta02\end{tabular}}}}%
    \put(-1.38003956,-0.56509841){\color[rgb]{0.34901961,0.34901961,0.34901961}\makebox(0,0)[lt]{\lineheight{1.25}\smash{\begin{tabular}[t]{l}Ta03\end{tabular}}}}%
    \put(-1.3266616,-0.56509841){\color[rgb]{0.34901961,0.34901961,0.34901961}\makebox(0,0)[lt]{\lineheight{1.25}\smash{\begin{tabular}[t]{l}Ta04\end{tabular}}}}%
    \put(-1.27328365,-0.56509841){\color[rgb]{0.34901961,0.34901961,0.34901961}\makebox(0,0)[lt]{\lineheight{1.25}\smash{\begin{tabular}[t]{l}Ta05\end{tabular}}}}%
    \put(-1.2199057,-0.56509841){\color[rgb]{0.34901961,0.34901961,0.34901961}\makebox(0,0)[lt]{\lineheight{1.25}\smash{\begin{tabular}[t]{l}Ta06\end{tabular}}}}%
    \put(-1.16652774,-0.56509841){\color[rgb]{0.34901961,0.34901961,0.34901961}\makebox(0,0)[lt]{\lineheight{1.25}\smash{\begin{tabular}[t]{l}Ta07\end{tabular}}}}%
    \put(-1.11314979,-0.56509841){\color[rgb]{0.34901961,0.34901961,0.34901961}\makebox(0,0)[lt]{\lineheight{1.25}\smash{\begin{tabular}[t]{l}Ta08\end{tabular}}}}%
    \put(-1.05977184,-0.56509841){\color[rgb]{0.34901961,0.34901961,0.34901961}\makebox(0,0)[lt]{\lineheight{1.25}\smash{\begin{tabular}[t]{l}Ta09\end{tabular}}}}%
    \put(-1.00639388,-0.56509841){\color[rgb]{0.34901961,0.34901961,0.34901961}\makebox(0,0)[lt]{\lineheight{1.25}\smash{\begin{tabular}[t]{l}Ta10\end{tabular}}}}%
    \put(-1.26475049,-0.59563438){\color[rgb]{0.34901961,0.34901961,0.34901961}\makebox(0,0)[lt]{\lineheight{1.25}\smash{\begin{tabular}[t]{l}Instance\end{tabular}}}}%
    \put(-1.37810108,-0.30089593){\color[rgb]{0.34901961,0.34901961,0.34901961}\makebox(0,0)[lt]{\lineheight{1.25}\smash{\begin{tabular}[t]{l}Performance over benchmark instances\end{tabular}}}}%
    \put(0,0){\includegraphics[width=\unitlength,page=2]{./figures/schedule.pdf}}%
    \put(-0.92031189,-0.3367425){\color[rgb]{0.34901961,0.34901961,0.34901961}\makebox(0,0)[lt]{\lineheight{1.25}\smash{\begin{tabular}[t]{l}Order Swap Mechanism\end{tabular}}}}%
    \put(0,0){\includegraphics[width=\unitlength,page=3]{./figures/schedule.pdf}}%
    \put(-0.92031189,-0.36506571){\color[rgb]{0.34901961,0.34901961,0.34901961}\makebox(0,0)[lt]{\lineheight{1.25}\smash{\begin{tabular}[t]{l}MWKR\end{tabular}}}}%
    \put(0,0){\includegraphics[width=\unitlength,page=4]{./figures/schedule.pdf}}%
    \put(-0.92031189,-0.39338893){\color[rgb]{0.34901961,0.34901961,0.34901961}\makebox(0,0)[lt]{\lineheight{1.25}\smash{\begin{tabular}[t]{l}SPT\end{tabular}}}}%
    \put(0,0){\includegraphics[width=\unitlength,page=5]{./figures/schedule.pdf}}%
    \put(-0.92031189,-0.42171214){\color[rgb]{0.34901961,0.34901961,0.34901961}\makebox(0,0)[lt]{\lineheight{1.25}\smash{\begin{tabular}[t]{l}Lower Bound\end{tabular}}}}%
    \put(0,0){\includegraphics[width=\unitlength,page=6]{./figures/schedule.pdf}}%
    \put(-0.0007065,0.23562364){\color[rgb]{0,0,0}\makebox(0,0)[lt]{\lineheight{1.25}\smash{\begin{tabular}[t]{l}machine 3\end{tabular}}}}%
    \put(-0.00013292,0.17023577){\color[rgb]{0,0,0}\makebox(0,0)[lt]{\lineheight{1.25}\smash{\begin{tabular}[t]{l}machine 2\end{tabular}}}}%
    \put(-0.00013292,0.10484791){\color[rgb]{0,0,0}\makebox(0,0)[lt]{\lineheight{1.25}\smash{\begin{tabular}[t]{l}machine 1\end{tabular}}}}%
    \put(0,0){\includegraphics[width=\unitlength,page=7]{./figures/schedule.pdf}}%
    \put(0.23504858,0.03804901){\color[rgb]{0,0,0}\makebox(0,0)[lt]{\lineheight{1.25}\smash{\begin{tabular}[t]{l}10\end{tabular}}}}%
    \put(0.36627075,0.03804901){\color[rgb]{0,0,0}\makebox(0,0)[lt]{\lineheight{1.25}\smash{\begin{tabular}[t]{l}20\end{tabular}}}}%
    \put(0.50601866,0.03804901){\color[rgb]{0,0,0}\makebox(0,0)[lt]{\lineheight{1.25}\smash{\begin{tabular}[t]{l}32\end{tabular}}}}%
    \put(0.59461216,0.03804901){\color[rgb]{0,0,0}\makebox(0,0)[lt]{\lineheight{1.25}\smash{\begin{tabular}[t]{l}37\end{tabular}}}}%
    \put(0.85668581,0.03804901){\color[rgb]{0,0,0}\makebox(0,0)[lt]{\lineheight{1.25}\smash{\begin{tabular}[t]{l}51\end{tabular}}}}%
    \put(0.46878246,0.0051339){\color[rgb]{0,0,0}\makebox(0,0)[lt]{\lineheight{1.25}\smash{\begin{tabular}[t]{l}time step\end{tabular}}}}%
    \put(0.94993039,0.2709144){\color[rgb]{0,0,0}\makebox(0,0)[lt]{\lineheight{1.25}\smash{\begin{tabular}[t]{l}job 1\end{tabular}}}}%
    \put(0.94993039,0.22272546){\color[rgb]{0,0,0}\makebox(0,0)[lt]{\lineheight{1.25}\smash{\begin{tabular}[t]{l}job 2\end{tabular}}}}%
    \put(0,0){\includegraphics[width=\unitlength,page=8]{./figures/schedule.pdf}}%
    \put(0.81850533,0.03804901){\color[rgb]{0,0,0}\makebox(0,0)[lt]{\lineheight{1.25}\smash{\begin{tabular}[t]{l}49\end{tabular}}}}%
    \put(0,0){\includegraphics[width=\unitlength,page=9]{./figures/schedule.pdf}}%
  \end{picture}%
\endgroup%

%% file: figures/comparison_left.pdf_tex
\begingroup%
  \makeatletter%
  \providecommand\color[2][]{%
    \errmessage{(Inkscape) Color is used for the text in Inkscape, but the package 'color.sty' is not loaded}%
    \renewcommand\color[2][]{}%
  }%
  \providecommand\transparent[1]{%
    \errmessage{(Inkscape) Transparency is used (non-zero) for the text in Inkscape, but the package 'transparent.sty' is not loaded}%
    \renewcommand\transparent[1]{}%
  }%
  \providecommand\rotatebox[2]{#2}%
  \newcommand*\fsize{\dimexpr\f@size pt\relax}%
  \newcommand*\lineheight[1]{\fontsize{\fsize}{#1\fsize}\selectfont}%
  \ifx\svgwidth\undefined%
    \setlength{\unitlength}{480bp}%
    \ifx\svgscale\undefined%
      \relax%
    \else%
      \setlength{\unitlength}{\unitlength * \real{\svgscale}}%
    \fi%
  \else%
    \setlength{\unitlength}{\svgwidth}%
  \fi%
  \global\let\svgwidth\undefined%
  \global\let\svgscale\undefined%
  \makeatother%
  \begin{picture}(1,0.58166718)%
    \lineheight{1}%
    \setlength\tabcolsep{0pt}%
    \put(0,0){\includegraphics[width=\unitlength,page=1]{figures/comparison_left.pdf}}%
    \put(0.78676329,0.20016334){\color[rgb]{0,0,0}\makebox(0,0)[lt]{\lineheight{1.25}\smash{\begin{tabular}[t]{l}\\conventional env +\\\ac{osm}\end{tabular}}}}%
    \put(0.78603251,0.20378307){\color[rgb]{0,0,0}\makebox(0,0)[lt]{\lineheight{1.25}\smash{\begin{tabular}[t]{l}conventional env\end{tabular}}}}%
    \put(0,0){\includegraphics[width=\unitlength,page=2]{figures/comparison_left.pdf}}%
    \put(0.03063532,0.52898881){\color[rgb]{0,0,0}\makebox(0,0)[lt]{\lineheight{0}\smash{\begin{tabular}[t]{l}$1.6\cdot 10^{3}$\end{tabular}}}}%
    \put(0.03063532,0.47550575){\color[rgb]{0,0,0}\makebox(0,0)[lt]{\lineheight{0}\smash{\begin{tabular}[t]{l}$1.4\cdot 10^{3}$\end{tabular}}}}%
    \put(0.03063532,0.4203962){\color[rgb]{0,0,0}\makebox(0,0)[lt]{\lineheight{0}\smash{\begin{tabular}[t]{l}$1.2\cdot 10^{3}$\end{tabular}}}}%
    \put(0.04344443,0.36204497){\color[rgb]{0,0,0}\makebox(0,0)[lt]{\lineheight{0}\smash{\begin{tabular}[t]{l}$1\cdot 10^{3}$\end{tabular}}}}%
    \put(0.46825799,0.01302009){\color[rgb]{0,0,0}\makebox(0,0)[lt]{\lineheight{0}\smash{\begin{tabular}[t]{l}training steps\end{tabular}}}}%
    \put(0.01603962,0.25047609){\color[rgb]{0,0,0}\rotatebox{90}{\makebox(0,0)[lt]{\lineheight{0}\smash{\begin{tabular}[t]{l}performance\end{tabular}}}}}%
    \put(0.05689393,0.31008972){\color[rgb]{0,0,0}\makebox(0,0)[lt]{\lineheight{0}\smash{\begin{tabular}[t]{l}$800$\end{tabular}}}}%
    \put(0.05689393,0.25552044){\color[rgb]{0,0,0}\makebox(0,0)[lt]{\lineheight{0}\smash{\begin{tabular}[t]{l}$600$\end{tabular}}}}%
    \put(0.05689393,0.19824974){\color[rgb]{0,0,0}\makebox(0,0)[lt]{\lineheight{0}\smash{\begin{tabular}[t]{l}$400$\end{tabular}}}}%
    \put(0.05689393,0.13935822){\color[rgb]{0,0,0}\makebox(0,0)[lt]{\lineheight{0}\smash{\begin{tabular}[t]{l}$200$\end{tabular}}}}%
    \put(0.07057564,0.05684041){\color[rgb]{0,0,0}\makebox(0,0)[lt]{\lineheight{0}\smash{\begin{tabular}[t]{l}$0$\end{tabular}}}}%
    \put(0.21096082,0.05346035){\color[rgb]{0,0,0}\makebox(0,0)[lt]{\lineheight{0}\smash{\begin{tabular}[t]{l}500k\end{tabular}}}}%
    \put(0.32843935,0.05314289){\color[rgb]{0,0,0}\makebox(0,0)[lt]{\lineheight{0}\smash{\begin{tabular}[t]{l}1M\end{tabular}}}}%
    \put(0.42777033,0.05346035){\color[rgb]{0,0,0}\makebox(0,0)[lt]{\lineheight{0}\smash{\begin{tabular}[t]{l}1.5M\end{tabular}}}}%
    \put(0.54467529,0.05314289){\color[rgb]{0,0,0}\makebox(0,0)[lt]{\lineheight{0}\smash{\begin{tabular}[t]{l}2M\end{tabular}}}}%
    \put(0.64171407,0.05346035){\color[rgb]{0,0,0}\makebox(0,0)[lt]{\lineheight{0}\smash{\begin{tabular}[t]{l}2.5M\end{tabular}}}}%
    \put(0.75670876,0.05346035){\color[rgb]{0,0,0}\makebox(0,0)[lt]{\lineheight{0}\smash{\begin{tabular}[t]{l}3M\end{tabular}}}}%
    \put(0.85489366,0.05346035){\color[rgb]{0,0,0}\makebox(0,0)[lt]{\lineheight{0}\smash{\begin{tabular}[t]{l}3.5M\end{tabular}}}}%
    \put(0.97103443,0.05314289){\color[rgb]{0,0,0}\makebox(0,0)[lt]{\lineheight{0}\smash{\begin{tabular}[t]{l}4M\end{tabular}}}}%
    \put(0,0){\includegraphics[width=\unitlength,page=3]{figures/comparison_left.pdf}}%
  \end{picture}%
\endgroup%

%% file: figures/comparison_right.pdf_tex
\begingroup%
  \makeatletter%
  \providecommand\color[2][]{%
    \errmessage{(Inkscape) Color is used for the text in Inkscape, but the package 'color.sty' is not loaded}%
    \renewcommand\color[2][]{}%
  }%
  \providecommand\transparent[1]{%
    \errmessage{(Inkscape) Transparency is used (non-zero) for the text in Inkscape, but the package 'transparent.sty' is not loaded}%
    \renewcommand\transparent[1]{}%
  }%
  \providecommand\rotatebox[2]{#2}%
  \newcommand*\fsize{\dimexpr\f@size pt\relax}%
  \newcommand*\lineheight[1]{\fontsize{\fsize}{#1\fsize}\selectfont}%
  \ifx\svgwidth\undefined%
    \setlength{\unitlength}{479.80305349bp}%
    \ifx\svgscale\undefined%
      \relax%
    \else%
      \setlength{\unitlength}{\unitlength * \real{\svgscale}}%
    \fi%
  \else%
    \setlength{\unitlength}{\svgwidth}%
  \fi%
  \global\let\svgwidth\undefined%
  \global\let\svgscale\undefined%
  \makeatother%
  \begin{picture}(1,0.56211761)%
    \lineheight{1}%
    \setlength\tabcolsep{0pt}%
    \put(0.07100025,0.08046833){\color[rgb]{0.12941176,0.12941176,0.12941176}\makebox(0,0)[rt]{\lineheight{1.25}\smash{\begin{tabular}[t]{r}100\end{tabular}}}}%
    \put(0.07100025,0.13745161){\color[rgb]{0.12941176,0.12941176,0.12941176}\makebox(0,0)[rt]{\lineheight{1.25}\smash{\begin{tabular}[t]{r}200\end{tabular}}}}%
    \put(0.07122868,0.1955773){\color[rgb]{0.12941176,0.12941176,0.12941176}\makebox(0,0)[rt]{\lineheight{1.25}\smash{\begin{tabular}[t]{r}300\end{tabular}}}}%
    \put(0.07191398,0.25507356){\color[rgb]{0.12941176,0.12941176,0.12941176}\makebox(0,0)[rt]{\lineheight{1.25}\smash{\begin{tabular}[t]{r}400\end{tabular}}}}%
    \put(0.07145712,0.3141128){\color[rgb]{0.12941176,0.12941176,0.12941176}\makebox(0,0)[rt]{\lineheight{1.25}\smash{\begin{tabular}[t]{r}500\end{tabular}}}}%
    \put(0.07077182,0.37109631){\color[rgb]{0.12941176,0.12941176,0.12941176}\makebox(0,0)[rt]{\lineheight{1.25}\smash{\begin{tabular}[t]{r}600\end{tabular}}}}%
    \put(0.07031496,0.43241983){\color[rgb]{0.12941176,0.12941176,0.12941176}\makebox(0,0)[rt]{\lineheight{1.25}\smash{\begin{tabular}[t]{r}700\end{tabular}}}}%
    \put(0.07077182,0.49054548){\color[rgb]{0.12941176,0.12941176,0.12941176}\makebox(0,0)[rt]{\lineheight{1.25}\smash{\begin{tabular}[t]{r}800\end{tabular}}}}%
    \put(0.07207944,0.54867127){\color[rgb]{0.12941176,0.12941176,0.12941176}\makebox(0,0)[rt]{\lineheight{1.25}\smash{\begin{tabular}[t]{r}900\end{tabular}}}}%
    \put(0,0){\includegraphics[width=\unitlength,page=1]{figures/comparison_right.pdf}}%
    \put(0.0677072,0.05799395){\color[rgb]{0.12941176,0.12941176,0.12941176}\makebox(0,0)[t]{\lineheight{1.25}\smash{\begin{tabular}[t]{c}0\end{tabular}}}}%
    \put(0.19984074,0.05724362){\color[rgb]{0.12941176,0.12941176,0.12941176}\makebox(0,0)[t]{\lineheight{1.25}\smash{\begin{tabular}[t]{c}500k\end{tabular}}}}%
    \put(0.31144302,0.05753322){\color[rgb]{0.12941176,0.12941176,0.12941176}\makebox(0,0)[t]{\lineheight{1.25}\smash{\begin{tabular}[t]{c}1M\end{tabular}}}}%
    \put(0.4268434,0.05753322){\color[rgb]{0.12941176,0.12941176,0.12941176}\makebox(0,0)[t]{\lineheight{1.25}\smash{\begin{tabular}[t]{c}1.5M\end{tabular}}}}%
    \put(0.53692615,0.05753322){\color[rgb]{0.12941176,0.12941176,0.12941176}\makebox(0,0)[t]{\lineheight{1.25}\smash{\begin{tabular}[t]{c}2M\end{tabular}}}}%
    \put(0.65080704,0.05753322){\color[rgb]{0.12941176,0.12941176,0.12941176}\makebox(0,0)[t]{\lineheight{1.25}\smash{\begin{tabular}[t]{c}2.5M\end{tabular}}}}%
    \put(0.76013046,0.05753322){\color[rgb]{0.12941176,0.12941176,0.12941176}\makebox(0,0)[t]{\lineheight{1.25}\smash{\begin{tabular}[t]{c}3M\end{tabular}}}}%
    \put(0.87629044,0.05753322){\color[rgb]{0.12941176,0.12941176,0.12941176}\makebox(0,0)[t]{\lineheight{1.25}\smash{\begin{tabular}[t]{c}3.5M\end{tabular}}}}%
    \put(0.98257531,0.05753322){\color[rgb]{0.12941176,0.12941176,0.12941176}\makebox(0,0)[t]{\lineheight{1.25}\smash{\begin{tabular}[t]{c}4M\end{tabular}}}}%
    \put(0,0){\includegraphics[width=\unitlength,page=2]{figures/comparison_right.pdf}}%
    \put(0.80134915,0.49170553){\color[rgb]{0,0,0}\makebox(0,0)[lt]{\lineheight{1.25}\smash{\begin{tabular}[t]{l}\\conventional env + \\\ac{osm}\end{tabular}}}}%
    \put(0.8035634,0.49856801){\color[rgb]{0,0,0}\makebox(0,0)[lt]{\lineheight{1.25}\smash{\begin{tabular}[t]{l}conventional env\end{tabular}}}}%
    \put(0,0){\includegraphics[width=\unitlength,page=3]{figures/comparison_right.pdf}}%
    \put(0.53483866,0.00555525){\color[rgb]{0.12941176,0.12941176,0.12941176}\makebox(0,0)[t]{\lineheight{1.25}\smash{\begin{tabular}[t]{c}training steps\end{tabular}}}}%
    \put(0.01870614,0.34371517){\color[rgb]{0.12941176,0.12941176,0.12941176}\rotatebox{90}{\makebox(0,0)[t]{\lineheight{1.25}\smash{\begin{tabular}[t]{c}episode length\end{tabular}}}}}%
    \put(0,0){\includegraphics[width=\unitlength,page=4]{figures/comparison_right.pdf}}%
  \end{picture}%
\endgroup%

%% file: figures/performance_left.pdf_tex
\begingroup%
  \makeatletter%
  \providecommand\color[2][]{%
    \errmessage{(Inkscape) Color is used for the text in Inkscape, but the package 'color.sty' is not loaded}%
    \renewcommand\color[2][]{}%
  }%
  \providecommand\transparent[1]{%
    \errmessage{(Inkscape) Transparency is used (non-zero) for the text in Inkscape, but the package 'transparent.sty' is not loaded}%
    \renewcommand\transparent[1]{}%
  }%
  \providecommand\rotatebox[2]{#2}%
  \newcommand*\fsize{\dimexpr\f@size pt\relax}%
  \newcommand*\lineheight[1]{\fontsize{\fsize}{#1\fsize}\selectfont}%
  \ifx\svgwidth\undefined%
    \setlength{\unitlength}{480bp}%
    \ifx\svgscale\undefined%
      \relax%
    \else%
      \setlength{\unitlength}{\unitlength * \real{\svgscale}}%
    \fi%
  \else%
    \setlength{\unitlength}{\svgwidth}%
  \fi%
  \global\let\svgwidth\undefined%
  \global\let\svgscale\undefined%
  \makeatother%
  \begin{picture}(1,0.63639193)%
    \lineheight{1}%
    \setlength\tabcolsep{0pt}%
    \put(0,0){\includegraphics[width=\unitlength,page=1]{./figures/performance_left.pdf}}%
    \put(0.00986522,0.62035083){\color[rgb]{0,0,0}\makebox(0,0)[lt]{\lineheight{1.25}\smash{\begin{tabular}[t]{l}$1.8 \cdot 10^{3}$\end{tabular}}}}%
    \put(0.00503661,0.56083308){\color[rgb]{0,0,0}\makebox(0,0)[lt]{\lineheight{1.25}\smash{\begin{tabular}[t]{l}$1.6 \cdot 10^{3}$\end{tabular}}}}%
    \put(0.00608115,0.49841736){\color[rgb]{0,0,0}\makebox(0,0)[lt]{\lineheight{1.25}\smash{\begin{tabular}[t]{l}$1.4 \cdot 10^{3}$\end{tabular}}}}%
    \put(0.00498967,0.43706261){\color[rgb]{0,0,0}\makebox(0,0)[lt]{\lineheight{1.25}\smash{\begin{tabular}[t]{l}$1.2 \cdot 10^{3}$\end{tabular}}}}%
    \put(0.02599587,0.37511807){\color[rgb]{0,0,0}\makebox(0,0)[lt]{\lineheight{1.25}\smash{\begin{tabular}[t]{l}$1 \cdot 10^{3}$\end{tabular}}}}%
    \put(0.03671347,0.31275163){\color[rgb]{0,0,0}\makebox(0,0)[lt]{\lineheight{1.25}\smash{\begin{tabular}[t]{l}$800$\end{tabular}}}}%
    \put(0.03826842,0.25005016){\color[rgb]{0,0,0}\makebox(0,0)[lt]{\lineheight{1.25}\smash{\begin{tabular}[t]{l}$600$\end{tabular}}}}%
    \put(0.83708847,0.17176422){\color[rgb]{0,0,0}\makebox(0,0)[lt]{\lineheight{1.25}\smash{\begin{tabular}[t]{l}$\tau = 0.00667$\end{tabular}}}}%
    \put(0.8360257,0.13545108){\color[rgb]{0,0,0}\makebox(0,0)[lt]{\lineheight{1.25}\smash{\begin{tabular}[t]{l}$\tau = 0.01$\end{tabular}}}}%
    \put(0.83565345,0.09703642){\color[rgb]{0,0,0}\makebox(0,0)[lt]{\lineheight{1.25}\smash{\begin{tabular}[t]{l}$T_p$\end{tabular}}}}%
    \put(0,0){\includegraphics[width=\unitlength,page=2]{./figures/performance_left.pdf}}%
    \put(0.06019101,0.04506179){\color[rgb]{0,0,0}\makebox(0,0)[lt]{\lineheight{1.25}\smash{\begin{tabular}[t]{l}0\end{tabular}}}}%
    \put(0.13642065,0.03293258){\color[rgb]{0,0,0}\makebox(0,0)[lt]{\lineheight{1.25}\smash{\begin{tabular}[t]{l}500k\end{tabular}}}}%
    \put(0.28193387,0.03589668){\color[rgb]{0,0,0}\makebox(0,0)[lt]{\lineheight{1.25}\smash{\begin{tabular}[t]{l}1.5M\end{tabular}}}}%
    \put(0.44302391,0.03531322){\color[rgb]{0,0,0}\makebox(0,0)[lt]{\lineheight{1.25}\smash{\begin{tabular}[t]{l}2.5M\end{tabular}}}}%
    \put(0.60134332,0.03648002){\color[rgb]{0,0,0}\makebox(0,0)[lt]{\lineheight{1.25}\smash{\begin{tabular}[t]{l}3.5M\end{tabular}}}}%
    \put(0.75044118,0.03688453){\color[rgb]{0,0,0}\makebox(0,0)[lt]{\lineheight{1.25}\smash{\begin{tabular}[t]{l}4.5M\end{tabular}}}}%
    \put(0.8995951,0.03589668){\color[rgb]{0,0,0}\makebox(0,0)[lt]{\lineheight{1.25}\smash{\begin{tabular}[t]{l}5.5M\end{tabular}}}}%
    \put(0.83825324,0.20803537){\color[rgb]{0,0,0}\makebox(0,0)[lt]{\lineheight{1.25}\smash{\begin{tabular}[t]{l}$\tau = 0.005$\end{tabular}}}}%
    \put(0,0){\includegraphics[width=\unitlength,page=3]{./figures/performance_left.pdf}}%
    \put(0.03886231,0.18922869){\color[rgb]{0,0,0}\makebox(0,0)[lt]{\lineheight{1.25}\smash{\begin{tabular}[t]{l}$400$\end{tabular}}}}%
    \put(0.0411345,0.12697283){\color[rgb]{0,0,0}\makebox(0,0)[lt]{\lineheight{1.25}\smash{\begin{tabular}[t]{l}$200$\end{tabular}}}}%
    \put(0.48267232,0.00695929){\color[rgb]{0,0,0}\makebox(0,0)[lt]{\lineheight{1.25}\smash{\begin{tabular}[t]{l}Training Steps\end{tabular}}}}%
    \put(0.01908312,0.30286929){\color[rgb]{0,0,0}\rotatebox{90}{\makebox(0,0)[lt]{\lineheight{1.25}\smash{\begin{tabular}[t]{l}Performance\end{tabular}}}}}%
    \put(0,0){\includegraphics[width=\unitlength,page=4]{./figures/performance_left.pdf}}%
  \end{picture}%
\endgroup%

%% file: figures/performance_right.pdf_tex
\begingroup%
  \makeatletter%
  \providecommand\color[2][]{%
    \errmessage{(Inkscape) Color is used for the text in Inkscape, but the package 'color.sty' is not loaded}%
    \renewcommand\color[2][]{}%
  }%
  \providecommand\transparent[1]{%
    \errmessage{(Inkscape) Transparency is used (non-zero) for the text in Inkscape, but the package 'transparent.sty' is not loaded}%
    \renewcommand\transparent[1]{}%
  }%
  \providecommand\rotatebox[2]{#2}%
  \newcommand*\fsize{\dimexpr\f@size pt\relax}%
  \newcommand*\lineheight[1]{\fontsize{\fsize}{#1\fsize}\selectfont}%
  \ifx\svgwidth\undefined%
    \setlength{\unitlength}{480bp}%
    \ifx\svgscale\undefined%
      \relax%
    \else%
      \setlength{\unitlength}{\unitlength * \real{\svgscale}}%
    \fi%
  \else%
    \setlength{\unitlength}{\svgwidth}%
  \fi%
  \global\let\svgwidth\undefined%
  \global\let\svgscale\undefined%
  \makeatother%
  \begin{picture}(1,0.56973748)%
    \lineheight{1}%
    \setlength\tabcolsep{0pt}%
    \put(0,0){\includegraphics[width=\unitlength,page=1]{./figures/performance_right.pdf}}%
    \put(0.01932409,0.52502598){\color[rgb]{0.12941176,0.12941176,0.12941176}\makebox(0,0)[lt]{\lineheight{0}\smash{\begin{tabular}[t]{l}800\end{tabular}}}}%
    \put(0.01975928,0.49750326){\color[rgb]{0.12941176,0.12941176,0.12941176}\makebox(0,0)[lt]{\lineheight{0}\smash{\begin{tabular}[t]{l}700\end{tabular}}}}%
    \put(0.01975928,0.54824846){\color[rgb]{0.12941176,0.12941176,0.12941176}\makebox(0,0)[lt]{\lineheight{0}\smash{\begin{tabular}[t]{l}900\end{tabular}}}}%
    \put(0.01938341,0.46761981){\color[rgb]{0.12941176,0.12941176,0.12941176}\makebox(0,0)[lt]{\lineheight{0}\smash{\begin{tabular}[t]{l}600\end{tabular}}}}%
    \put(0.02069903,0.43059468){\color[rgb]{0.12941176,0.12941176,0.12941176}\makebox(0,0)[lt]{\lineheight{0}\smash{\begin{tabular}[t]{l}500\end{tabular}}}}%
    \put(0.01994722,0.38417216){\color[rgb]{0.12941176,0.12941176,0.12941176}\makebox(0,0)[lt]{\lineheight{0}\smash{\begin{tabular}[t]{l}400\end{tabular}}}}%
    \put(0.02030237,0.32647659){\color[rgb]{0.12941176,0.12941176,0.12941176}\makebox(0,0)[lt]{\lineheight{0}\smash{\begin{tabular}[t]{l}300\end{tabular}}}}%
    \put(0.02155855,0.2439685){\color[rgb]{0.12941176,0.12941176,0.12941176}\makebox(0,0)[lt]{\lineheight{0}\smash{\begin{tabular}[t]{l}200\end{tabular}}}}%
    \put(0.02143002,0.10357334){\color[rgb]{0.12941176,0.12941176,0.12941176}\makebox(0,0)[lt]{\lineheight{0}\smash{\begin{tabular}[t]{l}100\end{tabular}}}}%
    \put(0.03087518,0.08240005){\color[rgb]{0.12941176,0.12941176,0.12941176}\makebox(0,0)[lt]{\lineheight{0}\smash{\begin{tabular}[t]{l}90\end{tabular}}}}%
    \put(0.04868432,0.04886346){\color[rgb]{0.12941176,0.12941176,0.12941176}\makebox(0,0)[lt]{\lineheight{0}\smash{\begin{tabular}[t]{l}0\end{tabular}}}}%
    \put(0.12470163,0.04354768){\color[rgb]{0.12941176,0.12941176,0.12941176}\makebox(0,0)[lt]{\lineheight{0}\smash{\begin{tabular}[t]{l}500k\end{tabular}}}}%
    \put(0.20178217,0.04248446){\color[rgb]{0.12941176,0.12941176,0.12941176}\makebox(0,0)[lt]{\lineheight{0}\smash{\begin{tabular}[t]{l}1M\end{tabular}}}}%
    \put(0.27789808,0.04299366){\color[rgb]{0.12941176,0.12941176,0.12941176}\makebox(0,0)[lt]{\lineheight{0}\smash{\begin{tabular}[t]{l}1.5M\end{tabular}}}}%
    \put(0.35215842,0.04364044){\color[rgb]{0.12941176,0.12941176,0.12941176}\makebox(0,0)[lt]{\lineheight{0}\smash{\begin{tabular}[t]{l}2M\end{tabular}}}}%
    \put(0.42498622,0.04470373){\color[rgb]{0.12941176,0.12941176,0.12941176}\makebox(0,0)[lt]{\lineheight{0}\smash{\begin{tabular}[t]{l}2.5M\end{tabular}}}}%
    \put(0.49994041,0.04576694){\color[rgb]{0.12941176,0.12941176,0.12941176}\makebox(0,0)[lt]{\lineheight{0}\smash{\begin{tabular}[t]{l}3M\end{tabular}}}}%
    \put(0.47074597,0.00314304){\color[rgb]{0.12941176,0.12941176,0.12941176}\makebox(0,0)[lt]{\lineheight{0}\smash{\begin{tabular}[t]{l}Training Steps\end{tabular}}}}%
    \put(0.01156096,0.28960531){\color[rgb]{0.12941176,0.12941176,0.12941176}\rotatebox{90}{\makebox(0,0)[lt]{\lineheight{0}\smash{\begin{tabular}[t]{l}Episode length\end{tabular}}}}}%
    \put(0.57223668,0.04523524){\color[rgb]{0.12941176,0.12941176,0.12941176}\makebox(0,0)[lt]{\lineheight{0}\smash{\begin{tabular}[t]{l}3.5M\end{tabular}}}}%
    \put(0.65463307,0.04576694){\color[rgb]{0.12941176,0.12941176,0.12941176}\makebox(0,0)[lt]{\lineheight{0}\smash{\begin{tabular}[t]{l}4M\end{tabular}}}}%
    \put(0.72267659,0.04576694){\color[rgb]{0.12941176,0.12941176,0.12941176}\makebox(0,0)[lt]{\lineheight{0}\smash{\begin{tabular}[t]{l}4.5M\end{tabular}}}}%
    \put(0.80135179,0.04576682){\color[rgb]{0.12941176,0.12941176,0.12941176}\makebox(0,0)[lt]{\lineheight{0}\smash{\begin{tabular}[t]{l}5M\end{tabular}}}}%
    \put(0.87258493,0.04523524){\color[rgb]{0.12941176,0.12941176,0.12941176}\makebox(0,0)[lt]{\lineheight{0}\smash{\begin{tabular}[t]{l}5.5M\end{tabular}}}}%
    \put(0.95391825,0.04629846){\color[rgb]{0.12941176,0.12941176,0.12941176}\makebox(0,0)[lt]{\lineheight{0}\smash{\begin{tabular}[t]{l}6M\end{tabular}}}}%
    \put(0.85593454,0.13850745){\color[rgb]{0,0,0}\makebox(0,0)[lt]{\lineheight{1.25}\smash{\begin{tabular}[t]{l}$\tau = 0.00667$\end{tabular}}}}%
    \put(0.85607791,0.10338089){\color[rgb]{0,0,0}\makebox(0,0)[lt]{\lineheight{1.25}\smash{\begin{tabular}[t]{l}$\tau = 0.01$\end{tabular}}}}%
    \put(0.8563986,0.17188802){\color[rgb]{0,0,0}\makebox(0,0)[lt]{\lineheight{1.25}\smash{\begin{tabular}[t]{l}$\tau = 0.005$\end{tabular}}}}%
    \put(0,0){\includegraphics[width=\unitlength,page=2]{./figures/performance_right.pdf}}%
  \end{picture}%
\endgroup%